\documentclass[10pt,twocolumn,letterpaper]{article}
\usepackage[pagenumbers]{cvpr}
\usepackage{tabularx}
\usepackage{multirow}
\usepackage{listings}

\lstdefinelanguage{json}{
    numbers=left,
    numberstyle=\small,
    frame=single,
    rulecolor=\color{black},
    showspaces=false,
    showtabs=false,
    breaklines=true,
    breakatwhitespace=true,
    basicstyle=\ttfamily\small,
    upquote=true,
    morestring=[b]",
    belowskip=0.5em,
    stringstyle=\color{string},
    morecomment=[l]{:},
    commentstyle=\color{keycolor},
    literate=
     *{0}{{{\color{numb}0}}}{1}
      {1}{{{\color{numb}1}}}{1}
      {2}{{{\color{numb}2}}}{1}
      {3}{{{\color{numb}3}}}{1}
      {4}{{{\color{numb}4}}}{1}
      {5}{{{\color{numb}5}}}{1}
      {6}{{{\color{numb}6}}}{1}
      {7}{{{\color{numb}7}}}{1}
      {8}{{{\color{numb}8}}}{1}
      {9}{{{\color{numb}9}}}{1}
      {\{}{{{\color{delim}{\{}}}}{1}
      {\}}{{{\color{delim}{\}}}}}{1}
      {[}{{{\color{delim}{[}}}}{1}
      {]}{{{\color{delim}{]}}}}{1},
}

\usepackage[dvipsnames]{xcolor}
\definecolor{delim}{RGB}{20,105,176}
\definecolor{string}{rgb}{0.64,0.08,0.08}
\definecolor{keycolor}{rgb}{0,0,1}
\definecolor{commentgreen}{rgb}{0,0.6,0}
\definecolor{codegreen}{rgb}{0,0.6,0}
\definecolor{codegray}{rgb}{0.5,0.5,0.5}
\definecolor{codepurple}{rgb}{0.58,0,0.82}
\definecolor{backcolour}{rgb}{0.95,0.95,0.92}

\definecolor{cvprblue}{rgb}{0.21,0.49,0.74}
\usepackage[pagebackref,breaklinks,colorlinks,citecolor=cvprblue]{hyperref}

\title{ModaVerse: Efficiently Transforming Modalities with LLMs}

\author{Xinyu Wang\\
University of Adelaide\\
\and
Bohan Zhuang\\
Monash University\\
\and
Qi Wu\\
University of Adelaide
}

\begin{document}

\twocolumn[{%
\renewcommand\twocolumn[1][]{#1}%
\maketitle
\begin{center}
\includegraphics[width=\textwidth]{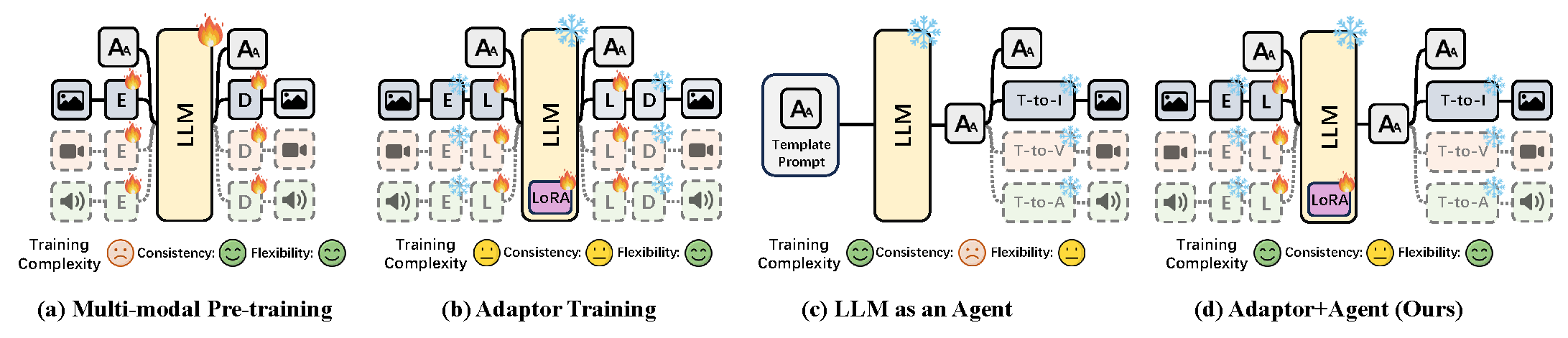}
\captionof{figure}{Comparative illustration of MLLM paradigms: (a) Multi-modal Pre-training, where new modules such as vision encoders and decoders are integrated within the standard LLM framework. (b)  Adaptor Training, illustrating the use of projection layers to connect LLMs to pre-existing modules. (c) LLM as an Agent, highlighting the strategic application of prompts in conjunction with external tools. (d) Adaptor+Agent (ours), transforming modalities with efficient language-based Input/Output (I/O) alignment. E, D, and L represent the Encoder, Decoder, and Linear Layer respectively. T-to-x denotes a text-to-x generative model, where x can be Image, Video, and Audio.
\label{fig:mmllm-type}}
\end{center}
}]

\begin{abstract}
Humans possess the capability to comprehend diverse modalities and seamlessly transfer information between them. In this work, we introduce ModaVerse, a Multi-modal Large Language Model (MLLM) capable of comprehending and transforming content across various modalities including images, videos, and audio. Predominant MLLM frameworks have largely relied on aligning latent spaces of textual and non-textual features. This alignment process, which synchronizes a language model trained on textual data with encoders and decoders trained on multi-modal data, often necessitates extensive training of several projection layers in multiple stages. Inspired by LLM-as-agent methodologies, we propose a novel Input/Output (I/O) alignment mechanism that operates directly at the level of natural language. It aligns the LLM's output with the input of generative models, avoiding the complexities associated with latent feature alignments, and simplifying the multiple training stages of existing MLLMs into a single, efficient process. By conducting experiments on several benchmarks, we demonstrate that our approach attains comparable performance with the state of the art while achieving considerable efficiencies in data usage. The code is available at \href{https://github.com/xinke-wang/ModaVerse}{https://github.com/xinke-wang/ModaVerse}.
\end{abstract}

\section{Introduction}

Spanning from ancient inscriptions to contemporary online encyclopedias, texts have served as the quintessential medium for chronicling the expanse of human knowledge. Such a vast accumulation of textual data provides a fertile terrain for training Large Language Models (LLMs)~\cite{radford2018improving, touvron2023llama, touvron2023llama2, radford2019language, brown2020language}. Through extensive training on massive corpora, LLMs undergo a transformative process, a phenomenon captured by the concept where quantitative increases result in qualitative behavioral shifts~\cite{wei2022emergent}, thus emerging with human-like reasoning abilities. This enables them to comprehend and respond to human instructions with remarkable precision. Such proficiency dramatically widens the scope of LLM applications across various domains, such as chatbots, programming copilots, and robotic agents.

Yet, the advent of richer communication forms calls for an evolution beyond the traditional confines of text. In the era where \textit{a picture is worth a thousand words}, the capability to interpret and integrate complex visual and auditory data is invaluable. The pursuit of enabling LLMs to process and generate information beyond textual data reflects the natural progression of AI, aspiring to mimic the full breadth of human communication. This has spurred the evolution of Multi-modal LLMs (MLLMs), which are designed to understand, transform, and produce content across various modalities, such as images, audio, and video. This growing interest has prompted a proliferation of research and innovation in the field~\cite{liu2023visual, yang2023dawn, zhu2023minigpt, wu2023next, koh2023generating, han2023imagebind, sun2023generative, yu2023scaling, wang2023switchgpt}. Addressing the limitations of traditional text-only LLMs, multi-modal pretraining, adaptor training, and LLM as an agent emerge as three key paradigms for equipping LLMs with multi-modal capabilities. Figure~\ref{fig:mmllm-type} compares the existing paradigm's overview schematic, assessing its performance and efficiency across three dimensions. Specifically, \textit{Training Complexity} refers to the volume of training data, computational resources consumed, and the number of training stages involved. \textit{Consistency} denotes the extent to which output is affected by modifications to inputs or prompts. \textit{Flexibility} pertains to the capacity for a variety of interpreting and generating outputs under diverse conditions.

\noindent\textbf{Multi-modal Pre-training} (Figure~\ref{fig:mmllm-type} (a)) expands the traditional LLM framework to accommodate non-textual inputs and outputs by integrating additional modality encoders and decoders into the existing framework. Through custom-designed pre-training tasks, the LLM learns to represent multiple modalities effectively, achieving superior consistency and flexibility compared to existing paradigms. However, adapting a text-based LLM, which has been pre-trained on extensive textual data, to a multi-modal context often requires significant fine-tuning or even complete retraining. Therefore, this adaptation demands considerable computational resources. For example, Emu~\cite{sun2023generative} combines Eva-CLIP~\cite{sun2023eva}, LLaMA~\cite{touvron2023llama}, and Stable Diffusion~\cite{rombach2022high} to develop a foundational multi-modal model. This development involves an intensive pre-training phase on a large-scale dataset and the use of hundreds of GPUs.

\noindent\textbf{Adaptor Training}  (Figure~\ref{fig:mmllm-type} (b)) offers a computationally economical alternative. This strategy, demonstrated by BLIP-2~\cite{Li2023ICML} and MiniGPT-4~\cite{zhu2023minigpt}, is typically built upon well-established LLMs and multi-modal encoders/decoders. It involves integrating these multi-modal modules with the LLM by training a set of projection layers while keeping the parameters of the LLM either frozen or fine-tuned using parameter-efficient techniques like LoRA~\cite{hu2021lora}. These layers translate non-textual representations, such as image features, into the textual domain of LLMs, thus avoiding extensive training but preserving flexibility. However, despite reducing training data volume and time, these methods still require a complex training procedure. For example, NExT-GPT~\cite{wu2023next} employs a three-step training pipeline where the encode/decode-side projection layers and the LLM adaptor are each trained in distinct stages. This intricate setup substantially escalates the complexity and leads to redundancy in the training process.

\noindent\textbf{LLM as an Agent} (Figure~\ref{fig:mmllm-type} (c)) demonstrates a training free framework. These methods utilize the zero-shot inference capabilities of LLMs, emphasizing strategic prompt crafting and workflow design. This approach guides the interpreting and generation of multi-modal content through interactions with external tools. For instance, HuggingGPT~\cite{shen2023hugginggpt} has developed a four-step pipeline that prompts OpenAI's ChatGPT to select and execute models from the HugingFace's model zoo, thereby solving a variety of tasks. However, it is crucial to recognize that these methods, lacking targeted training, often rely on x-to-text or text-x models, for processing non-textual inputs. This reliance may result in limited flexibility in handling diverse data types. Additionally, the heavy dependence on the design of system prompts and the reasoning capabilities of LLMs can further lead to inconsistent results.

So far, each paradigm presents a specialized approach for achieving functionality in MLLMs, each with its advantages and limitations. Considering these trade-offs, exploring the integration of their strengths into a cohesive approach is compelling. Specifically, this paper proposes Adaptor+Agent, an approach that aims to find a harmonious balance between the efficiencies of the LLM-as-agent approaches and the flexibility of adaptor training methods.

\begin{figure*}[t!]
    \centering
    \includegraphics[width=\linewidth]{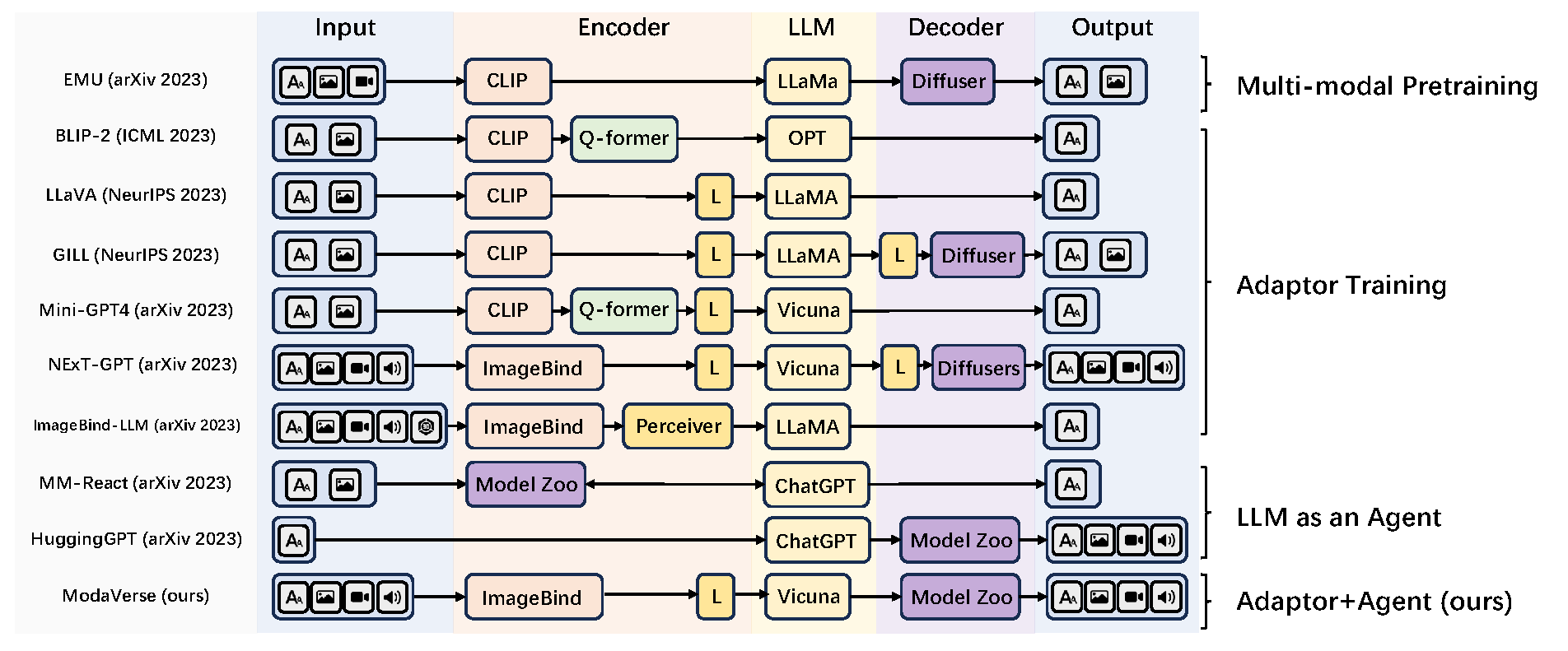}
    \caption{Comparison of the overview schematic of recent proposed MLLMs. L represents linear projection layers.}
    \label{fig:recent-mllm}
\end{figure*}

\noindent\textbf{Adaptor+Agent} (Figure~\ref{fig:mmllm-type} (d)) aims to combine the benefits of adaptor training with LLM-as-agent methods. As shown in the figure, to maintain the flexibility of accepting arbitrary combinations of input modalities, we train a set of linear adaptors to map the input's non-textual features into the LLM's textual space. This approach allows the model to comprehend multi-modal inputs while preserving training efficiency by only tuning the adaptors. For the output, we adopt an LLM-as-agent design, using established text-to-x models for generating non-text outputs. This strategy avoids the need for tuning additional output-side projection layers, thus enhancing efficiency. The primary challenge in the Adaptor+Agent framework is \textit{aligning the LLM's output with the text-to-x models' input}. To address this, we introduce Input/Output (I/O) Alignment. In contrast to previous adaptor-based approaches that focus on feature-level alignment between the LLM and generative models, our I/O Alignment strategy prompts the LLM to generate language-aligned meta-responses. These meta-responses contain detailed instructions for activating the generative models. We achieve this I/O Alignment through an instruction-following tuning process. As a result, in a single stage of tuning, the LLM is equipped to invoke external models for producing non-text outputs, thus bypassing the complex feature-level alignment typically required in the adaptor training paradigm.

In summary, the technical contributions of this paper are:

\begin{itemize}
    \item We introduce a new Adaptor+Agent training paradigm for Multi-modal Large Language Models that synthesizes the strengths of both adaptor training and the LLM-as-Agent approach. This integration effectively reaps the benefits of training efficiency and model flexibility.
    \item To address the alignment challenges inherent in the LLM-as-Agent methodology, we propose an I/O Alignment strategy. This strategy diverges from conventional feature-level alignment and instead operates at the natural language level, offering a more efficient alternative.
    \item Our final product, ModaVerse, demonstrates comparable performance to the current state of the arts on several widely used benchmarks while requiring fewer data and training resources, thereby offering a more efficient option without compromising effectiveness.
\end{itemize}

\begin{figure*}[t!]
    \centering
    \includegraphics[width=0.6\linewidth]{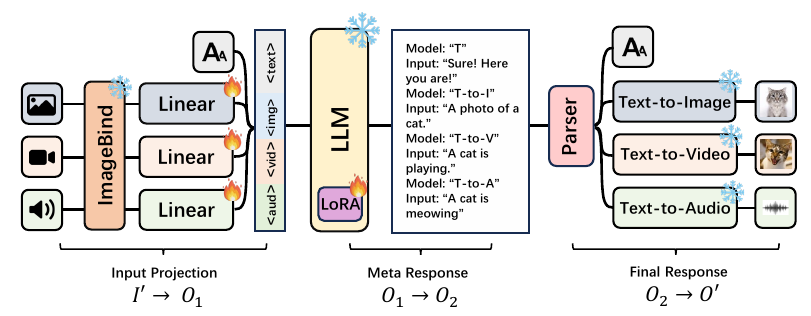}
    \caption{Overview of the Proposed ModaVerse Pipeline. In the input projection stage, multi-modal inputs $I'$ are aligned to the LLM's space $O_{1}$ using a series of trainable linear layers. During the meta-response generation stage, LLM is fine-tuned with a LoRA adaptor, prompting the generation of a meta-response $O_{2}$. In the final response generation stage, additional pretrained text-to-x models are utilized to generate the ultimate multi-modal response $O'$ based on the parsed meta response.}
    \label{fig:pipeline}
\end{figure*}

\section{Related Work}

\noindent\textbf{Multi-modal Pretrained MLLM.} Multi-modal pertaining is not a novel concept. Early efforts~\cite{su2019vl, li2020oscar, lu2019vilbert} have explored ways to extend the capabilities of language models to comprehend visual content. These models achieved promising performance on specific vision-language tasks~\cite{antol2015vqa} but failed to generalize to more general scenarios. Recent advancements, however, have revealed that simpler model architectures can yield impressive outcomes when subjected to extensive large-scale pretraining, thanks to advanced computational resources and diverse datasets. For example, PaLI~\cite{chen2022pali} demonstrates this by integrating a vision transformer with a language transformer and training on an extensive dataset of 10 billion image-text pairs. Similarly, CM3Leon~\cite{yu2023scaling} employs a straightforward decoder-only transformer architecture, trained on 340 million image-text pairs. This approach has enabled remarkable flexibility in generating and modifying both text and images, showing strong performance in image-to-text and text-to-image conversions. In addition, to enable an LLM that can generate non-text content, Emu~\cite{sun2023generative} combines a stable diffusion model with the LLaMA as a decoder, trained on a diverse corpus of 82 million image-text and video-text pairs. This integration marks a significant stride in the field, showcasing the growing versatility of LLM in multi-modal contexts.

\noindent\textbf{Adaptor Trained MLLM:} Leveraging recent advancements in parameter-efficient fine-tuning techniques~\cite{hu2021lora} and data-efficient approaches~\cite{liu2023visual}, numerous studies have explored the feasibility of training adaptors for aligning features between LLMs and various non-textual modules. Flamingo~\cite{alayrac2022flamingo} represents a pioneering effort in freezing the parameters of both visual encoders and LLMs, training a set of gated cross-attention layers to integrate visual knowledge into LLMs. However, it still necessitates extensive training on a massive dataset. Another notable example is BLIP-2~\cite{li2023blip}, which introduces a BERT-based Q-Former to translate image features into textual representations, thereby enabling LLMs to comprehend image content. This innovation has inspired subsequent research~\cite{su2023pandagpt, chen2023minigpt}, revealing that the Q-Former structure can be further simplified to a single linear layer, significantly reducing the number of trainable parameters. However, these advances, while showing considerable promise, have been predominantly applied to image-text tasks. In an effort to create a more versatile MLLM that can process a broader array of input types, subsequent works~\cite{su2023pandagpt, han2023imagebind} replaced the dual-modality encoder, CLIP~\cite{radford2021learning}, with the more versatile six-modality encoder, ImageBind~\cite{han2023imagebind}. In addition, other works~\cite{koh2023generating, wu2023next, dong2023dreamllm, zhang2023internlm} try to align the output side of LLMs with generation models, enabling the utilization of latent features from LLMs to guide generative models in producing non-textual content. For example, a concurrent study, NExT-GPT~\cite{wu2023next}, introduces a multi-stage training procedure. This procedure includes a series of adaptors that align the LLM with encoders and generative models at the feature level. Our work differs from NExT-GPT in that it aligns the generative models at the language level instead of the feature level, thereby significantly reducing training complexity.

\noindent\textbf{LLM-as-agent MLLM:} The remarkable zero-shot inference capabilities of LLMs enable them to effectively utilize external tools~\cite{schick2023toolformer, suris2023vipergpt, yang2023mm, lu2023chameleon, shen2023hugginggpt}. This potential facilitates the creation of specialized pipelines and associated prompts, guiding LLMs to understand or produce multi-modal content. For instance, HuggingGPT~\cite{openai2023chatgpt} has developed a multi-step pipeline. In this process, ChatGPT initially interprets human instructions and selects appropriate models from a model zoo to accomplish the given tasks. Subsequently, the outputs from these external models are fed back into ChatGPT for parsing and generating the final response. Another notable example is MM-React~\cite{yang2023mm}, which introduces the integration of vision experts, such as OCR, image captioning, and object detection models, to extend the LLMs' ability to process visual content. For each pair of input images and instructions, the LLM employs the relevant vision expert to extract pertinent information from the images, thereafter generating relevant responses.

Figure~\ref{fig:recent-mllm} presents a schematic overview of recently proposed MLLMs. It demonstrates the characteristic of adaptor training, wherein all models incorporate additional projection components, such as a linear layer or a Q-former, either before or after the LLM. These components are utilized to align textual and non-textual representations between the LLM and encoders/decoders. In contrast, multimodal pretraining methods usually feature a more straightforward and concise architecture. They direct the LLM itself to learn multimodal features, thus avoiding the projection structures between different modules. Furthermore, LLM-as-agent methods employ an external model zoo to assist in processing or producing non-textual content, without integrating trainable modules into the system. In comparison, the proposed Adaptor+Agent paradigm follows an adaptor structure on the encoder side, where linear projection layers are trained to align the input features with the LLM's textual space. On the decoder side, the LLM is treated as an agent to invoke external models for generating non-text content.

\section{ModaVerse}

\subsection{Pipeline Overview}
\label{sec:pipeline}

Figure~\ref{fig:pipeline} illustrates the comprehensive framework of the proposed ModaVerse, which contains three functional blocks, including input projection, meta-response generation, and final response generation.

\noindent\textbf{Input Projection:} To adapt a text-based LLM into an MLLM capable of interpreting non-textual inputs, it is essential to align the LLM's textual features with various modalities during the input phase. Recent research~\cite{zhu2023minigpt, chen2023minigpt, su2023pandagpt} has demonstrated the feasibility of aligning these different modalities using a single linear layer. Following this, we employ ImageBind~\cite{han2023imagebind} as a unified encoder, which processes inputs from diverse data types, including images, videos, and audio, converting them into a specific embedding. Subsequently, for each modality, we learn a set of linear projection layers to map these encoded representations into the LLM's text space. As a result, ModaVerse gains the capability to comprehend multi-modal inputs. 

\noindent\textbf{Meta Response Generation:} Since the foundational LLM is pre-trained exclusively on text-only data, it lacks the capability to directly generate non-text outputs. To address this limitation, we treat the foundational LLM as an agent, designed to produce only meta-responses. As depicted in Figure~\ref{fig:pipeline}, the meta-response comprises formatted information that includes the invocation details. For instance, according to the meta-response, the system might activate a text-to-image model to create an image based on the prompt ``A photo of a cat". This design circumvents the need for training an additional output-side projection layer to align the LLM's feature space with that of generative models, thereby simplifying the training process.

\noindent\textbf{Final Response Generation:} This block incorporates several replaceable text-to-x models to generate the final response, which may include images~\cite{rombach2022high}, videos~\cite{VideoFusion}, and audio~\cite{liu2023audioldm}. Based on the invocation details parsed from the meta-responses, one or more models will be activated to produce the non-textual output.

So far, the Adaptor+Agent paradigm has become clear. In this paradigm, the input projection is designed with a set of linear adaptors that map multimodal features to the LLM's textual space. The LLM itself is treated as an agent, invoking external models to generate the final responses. The benefits of such a design are:

\begin{itemize}
    \item Training adaptors during the input phase preserve the details in the input data, while simultaneously reducing the training volume compared to full multimodal pretraining.
    \item Treating the LLM as an agent during the output phase not only decouples it from external generative models, enabling a plug-and-play approach but also eliminates the need for additional projection layers. This means that running generative models during the training stage is unnecessary, thereby reducing training complexity.
\end{itemize}

\subsection{I/O Alignment}
\label{sec:io_alignment}

Consider a text-based LLM: $I\rightarrow O$ with its input and output sets defined as $I = O =\{text\}$, the objective of ModaVerse is to discover an efficient transformation that extends LLM to a multimodal model capable of handling $I' = O' =\{text, image, video, audio\}$, described as follows:

\begin{equation}
\left\{
\begin{array}{ll}
    P: \texttt{ImageBind}(I') \rightarrow O_{1}, \\
    LLM': O_{1} \rightarrow O_{2},\\
    M: O_{2} \rightarrow O'
\end{array} 
\right.
\end{equation}

\noindent Each line of this equation corresponds to a stage depicted in Figure~\ref{fig:pipeline}. The first line denotes a trainable projection $P$ from the multimodal feature - extracted by ImageBind~\cite{han2023imagebind} from the input $I'$ - into the textual space of the LLM, $O_{1}$. The second line involves tuning a LoRA~\cite{hu2021lora} adaptor to prompt the adapted LLM, defined as $LLM'$, to generate a meta-response $O_{2}$ from the input feature $O_{1}$. Finally, the third line, where $M$ represents the frozen, established text-to-x model zoo, utilizes the parsed meta-response to generate the final multimodal outputs $O'$. To achieve these objectives, we propose an instruction-following I/O Alignment. This alignment aims to simultaneously fit the $I'\rightarrow O_{1}$ and $O_{1}\rightarrow O_{2}$ alignments. As such, the trainable components, as depicted in Figure~\ref{fig:pipeline}, consist of three linear layers and the LoRA adaptor. Specifically, the I/O Alignment involves two primary components: the construction of instructions, and the tuning of linear and LoRA adaptors.

To implement I/O Alignment, two issues must be addressed. First, since an exact representation of $O_{1}$ is not directly obtainable, existing adaptor-based methods typically train a direct mapping from $I' \rightarrow O_{2}$, using captions from paired datasets as $O_{2}$ to learn the projections, framing the process as a multi-modal captioning task. However, in our case, $O_{2}$ is a meta response rather than just the text descriptions of $I'$. That is, given instructions and accompanying multi-modal inputs, the expected output should specifically identify \textit{which} model to use and \textit{how} to use it. For example, given the instruction, `Generate an image for an animal based on the provided audio clip of its vocalization', along with an accompanying audio clip that records a cat's meowing, the expected invocation information should be as follows: `\{model: ``text-to-image", prompts: ``a photo of a cat"\}'. Therefore, simply using x-to-text datasets to train the projection layers under an image captioning task is not possible to facilitate such purposes. The second issue is the language-level misalignment due to the different training corpora of LLMs and generative models. For instance, to describe a landscape image, an LLM tends to produce coherent, literary paragraphs, whereas a text-to-image model typically prefers concise, descriptive prompts accompanied by attributes such as ``4k" and ``masterpiece". Thus, I/O Alignment should also achieve $O_{1} \rightarrow O_{2}$, ensuring language-level alignment between the meta-response and the input prompts required by generative models.

\begin{table*}[t!]
\centering
\begin{tabular}{c|ccc|p{0.5cm}c|p{0.5cm}c|p{0.5cm}c}\hline
\multirow{2}{*}{Method} & \multirow{2}{*}{Type} & \multirow{2}{*}{Input} & \multirow{2}{*}{Output} & \multicolumn{2}{c|}{Stage I} & \multicolumn{2}{c|}{Stage II} & \multicolumn{2}{c}{Stage III} \\
 & & & & Data & GPU Time & Data & GPU Time & Data & GPU Time \\ \hline
 Emu~\cite{sun2023generative}& pretrain & T,I,V & T,I & 82M & 128$\times$2d & 24M & 32$\times$ & 1.28M & 16$\times$16h \\
 LLaVA~\cite{liu2023visual} & adaptor & T,I & T & 595k & 8$\times$20h & 158k & 8$\times$10h & N/A & N/A\\
 BLIP-2~\cite{li2023blip} & adaptor & T,I & T & 129M & 16$\times$6d & 129M & 16$\times$3d & N/A & N/A \\
 NExT-GPT~\cite{wu2023next} & adaptor & T,I,V,A & T,I,V,A & 13M+ & - & 13M+ & - & 20k+ & - \\
 ModaVerse & adaptor+agent & T,I,V,A & T,I,V,A & 2M & 4$\times$20h & N/A & N/A & N/A & N/A \\\hline  
\end{tabular}
\caption{Comparison of training complexity of ModaVerse with recent MLLMs. `N/A' indicates stages that are not required, while `-' denotes that the data was not disclosed by the authors. `T', `I', `V', `A' are the abbreviations of text, image, video, and audio, respectively.}
\label{table:training_complexity}
\end{table*}
 
To address the issues mentioned above, I/O Alignment employs an instruction-following training approach. The adaptors are trained with an input that includes both a language instruction and accompanying multi-modal elements, with the aim to produce a meta-response that details the following invocation. We utilize training datasets from generative models to create pairs of instructions and their corresponding ground truths (see Section~\ref{sec:instruction_generation} for the instruction generation procedure). This method is beneficial for several reasons: 1. Instruction-following tuning compels the LLM to fully comprehend multi-modal inputs, thereby aiding in aligning the input projection layers between multi-modal input and LLM. 2. The training datasets of generative models typically provide both non-textual data, such as images, videos, or audio, and their corresponding textual descriptions, thereby offering a solid foundation of aligned data samples. 3. Most open-source generative models are trained on the same publicly available datasets. Aligning the meta-response with the text descriptions from these datasets means it is possible to seamlessly switch between generative models, thus facilitating a plug-and-play approach. 

Based on this, we generate different types of instructions to fulfill the above objectives, which are as follows:

\noindent\textbf{Input-side Alignment Instruction} focuses on aligning the LLM's capability to comprehend inputs comprising combinations of various modal data, such as text+image, image+video, or image+audio+video. For instance, when presented with a combination of image, audio, and video inputs, the instruction ``Describe the given image, audio, and video" directs the MLLM to sequentially describe the content in the image, followed by the audio, and then the video.

\noindent\textbf{Output-side Alignment Instruction} aims to align the LLM's ability to generate meta-responses that include invocation details, such as the selected model and prompts. For example, the instruction ``Generate an image based on the provided audio of an animal sound" teaches the model to utilize a text-to-image model to generate an image, potentially with a prompt like ``A photo of a cat".

\noindent\textbf{Reasoning Boosting Instruction} is designed to preserve and enhance the LLM's reasoning capabilities through a diverse range of topics. For instance, an instruction like ``Where might this audio clip have been recorded?" requires the LLM to make a reasoned inference based on the input data, thereby strengthening its reasoning skills.

\subsection{Instruction Generation and Training}
\label{sec:instruction_generation}

In this section, we introduce how to generate the instruction invocation pairs used in I/O Alignment. 

\begin{lstlisting}[language=json, label=lst:validation_instruction]
{"instruction": ["Generate an image of an animal based on the provided vocalization.", "cat_meowing.wav", ]
"invocation": [("text-to-image", "A photo of a cat"), ]}
\end{lstlisting}

As demonstrated in the code block above, an instruction-invocation pair consists of two parts: the instruction, which represents the input, and the invocation, which represents the expected output of the LLM. To obtain these forms of data samples, we constructed them from two sources. The first source utilizes components of existing instruction tuning works, such as LLaVA~\cite{liu2023visual}, VideoChat~\cite{li2023videochat}, and InstructBLIP~\cite{liu2023improved}. However, it should be noted that the instructions in these works primarily fall into two categories: Input-side Alignment Instructions and Reasoning Boosting Instructions. Thus, to generate Output-side Alignment Instructions, which are crucial for the success of ModaVerse, we created specific templates to assist the OpenAI ChatGPT API in producing new instructions on a large scale. Specifically, each query sent to the ChatGPT API comprises three components: Seed Examples, Candidate Descriptions, and Language References.

\noindent\textbf{Seed Examples} consist of a set of standard instruction invocation pairs, randomly selected from a manually crafted collection. These examples serve as guides for the ChatGPT API, demonstrating how to generate samples in the given format and providing an illustration of the task.

\noindent\textbf{Candidate Descriptions} comprise randomly selected text descriptions from the paired datasets, which include descriptions of images, audio, and videos. These descriptions aim to mimic the true inputs, while the ChatGPT API is requested to generate appropriate instructions and invocation details based on these candidates and seed examples.

\noindent\textbf{Language References} include text descriptions randomly selected from the training set of generative models. These samples serve as a guide for the ChatGPT API to learn the language style of the prompts used in the generative models, helping to generate language-aligned invocation details.

\begin{table*}[!t]
\begin{minipage}{\textwidth}
\centering
\begin{minipage}[b]{0.48\textwidth}
\fontsize{8}{11}\selectfont
\setlength{\tabcolsep}{1.5mm}
\centering
\begin{tabular}{lc}
\hline
\bf Method & \bf FID ($\downarrow$) \\
\hline
CogVideo~\cite{ding2021cogview} (NeurIPS'21) & 27.10 \\   
GLIDE~\cite{nichol2021glide} (ICML'22) & 	12.24 \\
CoDi~\cite{tang2023any} (NeurIPS'23) &  11.26 \\
SD~\cite{rombach2022high} (CVPR'22) & 11.21 \\
NExT-GPT~\cite{wu2023next} (arXiv'23) & 11.28\\\hline
ModaVerse & 11.24 \\\hline
\end{tabular}
\captionof{table}{
\label{tab:T2I-res}
Text-to-image performance on COCO-caption~\cite{lin2014microsoft}.
}
\end{minipage}
\hfill
\begin{minipage}[b]{0.48\textwidth}
\fontsize{8}{11}\selectfont
\setlength{\tabcolsep}{1.2mm}
\centering
\begin{tabular}{lccc}
\hline
\bf Method & \bf B@4 ($\uparrow$) & \bf METEOR ($\uparrow$) & \bf CIDEr ($\uparrow$)\\
\hline
Oscar~\cite{li2020oscar} (ECCV'20) & 36.58 & 30.4 & 124.12\\
BLIP-2~\cite{Li2023ICML} (ICML'23) & 43.7 & --- & 145.8\\
OFA~\cite{WangYMLBLMZZY22} (ICML'22) & 44.9 & 32.5 & 154.9\\
CoDi~\cite{tang2023any} (NeurIPS'23) & 40.2 & 31.0 & 149.9\\
NExT-GPT~\cite{wu2023next} (arXiv'23) & 44.3 & 32.9 & 156.7\\\hline
ModaVerse & 43.9 & 31.8 & 151.4\\\hline
\end{tabular}
\captionof{table}{
\label{tab:I2T-res}
Image-to-text performance on COCO-caption data \cite{lin2014microsoft}.
} 
\end{minipage}
\end{minipage}
\end{table*}

\begin{table*}[!t]
\begin{minipage}{\textwidth}
\centering
\begin{minipage}[b]{0.48\textwidth}
\centering
\fontsize{8}{11}\selectfont
\setlength{\tabcolsep}{1.5mm}
\begin{tabular}{lcc}
\hline
\bf Method & \bf FD ($\downarrow$) & \bf IS ($\uparrow$) \\
\hline
DiffSound~\cite{YangYWWWZY23} (TASLP'23)&	47.68 & 4.01 \\   
AudioLDM-S~\cite{liu2023audioldm} (ICML'23) & 29.48 & 6.90 \\   
AudioLDM-L~\cite{liu2023audioldm} (ICML'23) &  23.31  &  8.13 \\   
CoDi~\cite{tang2023any} (NeurIPS'23) & 22.90 & 8.77 \\
NExT-GPT~\cite{wu2023next} (arXiv'23) & 23.58 & 8.35 \\\hline
ModaVerse & 23.40 & 8.22 \\\hline
\end{tabular}
\captionof{table}{
\label{tab:T2A-res}
Text-to-audio performance on AudioCaps \cite{KimKLK19}.
}
\end{minipage}
\hfill
\begin{minipage}[b]{0.48\textwidth}
\centering
\fontsize{8}{11}\selectfont
\setlength{\tabcolsep}{1.1mm}
\begin{tabular}{lcc}
\hline
\bf Method & \bf SPIDEr ($\uparrow$) & \bf CIDEr ($\uparrow$)\\
\hline
AudioCaps~\cite{KimKLK19} (NAACL'19) &	 0.369& 0.593 \\   
BART~\cite{gontier2021automated} (DCASE'21) &  0.465& 0.753 \\   
AL-MixGen~\cite{kim2022improving} (ArXiv'22) &  0.466 & 0.755 \\   
CoDi~\cite{tang2023any} (NeurIPS'23) & 0.480 & 0.789 \\
NExT-GPT~\cite{wu2023next} (arXiv'23) & 0.521 & 0.802 \\\hline
ModaVerse & 0.494 & 0.792 \\\hline
\end{tabular}
\captionof{table}{
\label{tab:A2T-res}
Audio-to-text performance on AudioCaps \cite{KimKLK19}.
}
\end{minipage}
\end{minipage}
\end{table*}

\begin{table*}[!t]
\begin{minipage}{\textwidth}
\centering
\begin{minipage}[b]{0.48\textwidth}
\centering
\fontsize{8}{11}\selectfont
\setlength{\tabcolsep}{1.1mm}
\begin{tabular}{lcc}
\hline
\bf Method & \bf FID ($\downarrow$) & \bf CLIPSIM ($\uparrow$) \\
\hline
CogVideo~\cite{hong2022cogvideo} (NeurIPS'21) &	23.59 & 	0.2631 \\   
MakeVideo~\cite{huang2023make} (ICML'23) & 	13.17 & 	0.3049 \\   
Latent-VDM~\cite{rombach2022high} (CVPR'22) &  14.25 & 	0.2756 \\   
Latent-Shift~\cite{an2023latent} (arXiv'23) & 15.23 & 	0.2773 \\
CoDi~\cite{tang2023any} (NeurIPS'23) & --- &  0.2890 \\
NExT-GPT~\cite{wu2023next} (arXiv'23) & 13.04 & 0.3085 \\\hline
ModaVerse & 13.35 & 0.3014 \\\hline
\end{tabular}
\captionof{table}{
\label{tab:T2V-res}
Text-to-video performance on MSR-VTT~\cite{xu2016msr}.
}
\end{minipage}
\hfill
\begin{minipage}[b]{0.45\textwidth}
\centering
\fontsize{8}{11}\selectfont
\setlength{\tabcolsep}{1.1mm}
\begin{tabular}{lcc}
\hline
\bf Method & \bf B@4 ($\uparrow$) & \bf METEOR ($\uparrow$)  \\
\hline
ORG-TRL~\cite{ZhangSY0WHZ20} (CVPR'20) & 43.6 & 28.8 \\   
GIT~\cite{WangYHLLGLLW22} (TMLR'22) & 54.8 & 33.1 \\   
mPLUG-2~\cite{xu2023mplug} (ICML'23) & 57.8 & 34.9 \\   
CoDi~\cite{tang2023any} (NeurIPS'23) & 52.1 & 32.5 \\
NExT-GPT~\cite{wu2023next} (arXiv'23) & 58.4 & 38.5 \\\hline
ModaVerse & 56.5 & 35.2 \\\hline
\end{tabular}
\captionof{table}{
\label{tab:V2T-res}
Video-to-text performance on MSR-VTT~\cite{xu2016msr}.
}
\end{minipage}
\end{minipage}
\end{table*}

For training, we use Vicuna~\cite{chiang2023vicuna} as the foundation LLM, the trainable parts of the proposed ModaVerse (see Figure~\ref{fig:pipeline}) consist only of three linear layers and the LoRA adaptor of the LLM. Together, these components comprise about 40M trainable parameters. Table~\ref{table:training_complexity} compares the training complexity of ModaVerse with that of some recently proposed MLLMs. It shows that the proposed method enjoys lower training complexity.

\begin{figure*}[t!]
    \centering
    \includegraphics[width=0.85\linewidth]{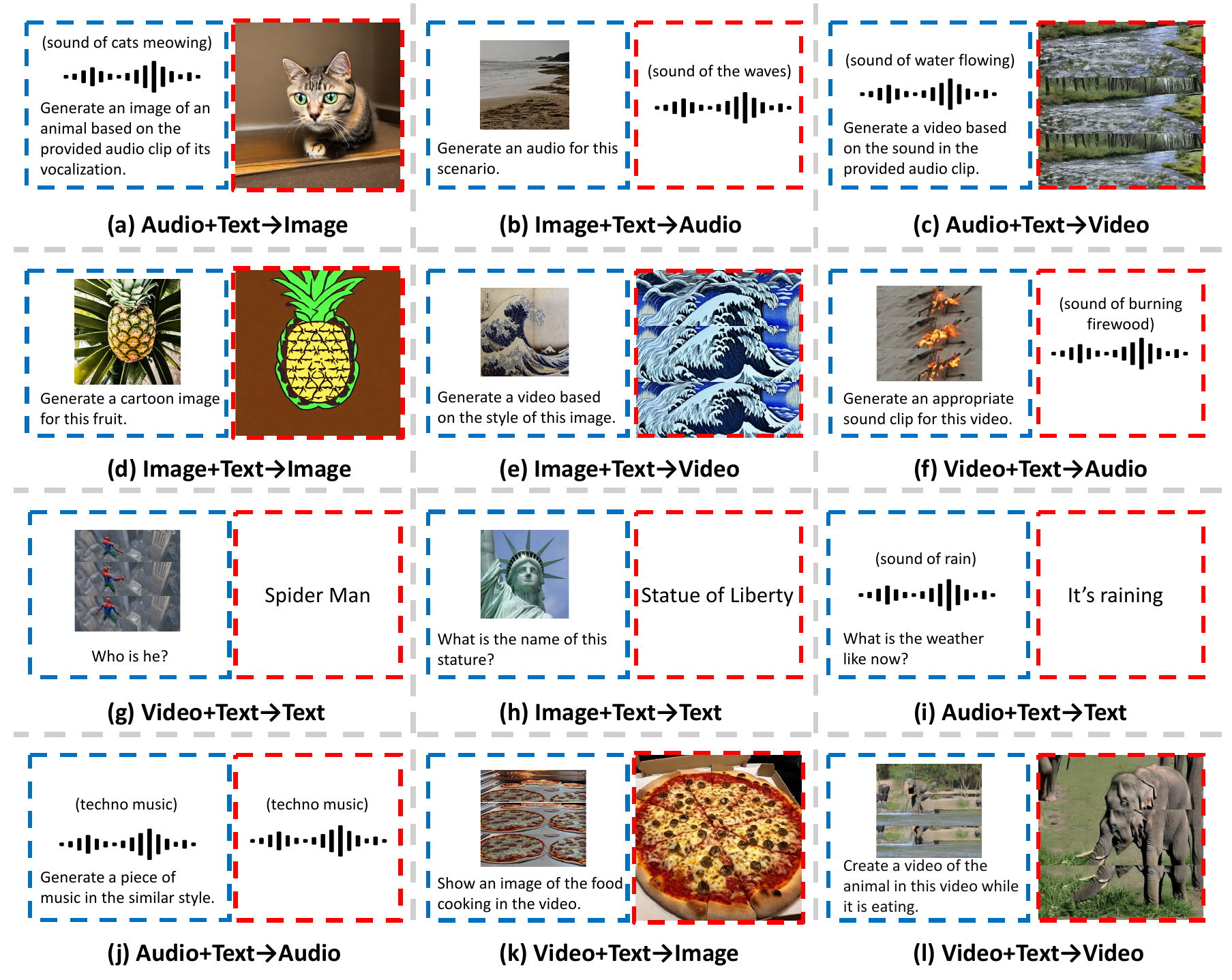}
    \caption{Qualitative examples of the proposed ModaVerse interpreting and producing data presented in combinations of various modalities. Blue and Red dashed boxes represent input and output respectively.}
    \label{fig:qualitative}
\end{figure*}

\section{Experiments}

\subsection{Quantitative Results}
\label{sec:quantitative}

To evaluate the proposed ModaVerse, we follow previous works~\cite{tang2023any} to assess the model's understanding ability (x$\rightarrow$text) and its generation ability (text$\rightarrow$x), where x can be image, audio, and video. Tables~\ref{tab:T2I-res} and~\ref{tab:I2T-res} illustrate the text-to-image and image-to-text performance on the COCO caption dataset~\cite{lin2014microsoft}. The results demonstrate that our method achieved an 11.28 FID score in the image generation task, comparable to recent methods. In terms of image understanding capability, the proposed ModaVerse outperforms the any-to-any diffuser CoDi~\cite{tang2023any} in both B@4 and METEOR metrics, though it is slightly lower than NExT-GPT~\cite{wu2023next} and OFA~\cite{WangYMLBLMZZY22}. The text-to-audio and audio-to-text performances on the AudioCaps dataset~\cite{KimKLK19} are showed in Tables~\ref{tab:T2A-res} and~\ref{tab:A2T-res}. The proposed method outperforms NExT-GPT and is narrowly eclipsed by CoDi. In audio captioning, our approach secures the second-best performance, rivaling the state-of-the-art models. Tables~\ref{tab:T2V-res} and~\ref{tab:V2T-res} showcase the text-to-video and video-to-text performances on MSR-VTT~\cite{xu2016msr}. Our method achieved a 13.35 FID score and a 0.3014 CLIPSIM score in video generation, demonstrating parity with top-tier methods. Similarly, our approach shows competitive results in video-to-text tasks, as evidenced by its B@4 and METEOR scores.

Although our method does not outperform all state-of-the-art methods (including two concurrent arXiv submissions) across the six benchmarks, it is important to note the efficiency of the proposed ModaVerse. First, our method is capable of converting a variety of modalities, whereas some state-of-the-art methods, such as SD~\cite{rombach2022high} and OFA~\cite{WangYMLBLMZZY22}, are specifically designed for single-route conversions like text-to-image. Second, as Table~\ref{table:training_complexity} illustrates, ModaVerse benefits from a more efficient training paradigm. It requires less data and fewer computational resources. Specifically, in contrast to NExT-GPT which necessitates three stages to independently train the projection layers, our method streamlines this process into a single stage. Regarding training data, our approach uses less than 2\% of the data volume required by Emu~\cite{koh2023generating} and BLIP-2~\cite{li2023blip}.

\subsection{Qualitative Results}

Since publicly available datasets are limited to certain common modality combinations, such as image-to-text and text-to-video, this limitation may not comprehensively capture the full extent of ModaVerse's capabilities. Therefore, Figure~\ref{fig:qualitative} showcases various qualitative results of ModaVerse across different modalities. For instance, examples (a), (c), (f), and (l) emphasize the model's conditioned generative abilities. In addition, examples (g), (h), and (i) demonstrate its proficiency in answering questions with inputs from a variety of modalities. Moreover, examples (d) and (e) demonstrate the potential for style transfer.

\subsection{Limitations and Failure Cases}

\begin{figure}
    \centering
    \includegraphics[width=\linewidth]{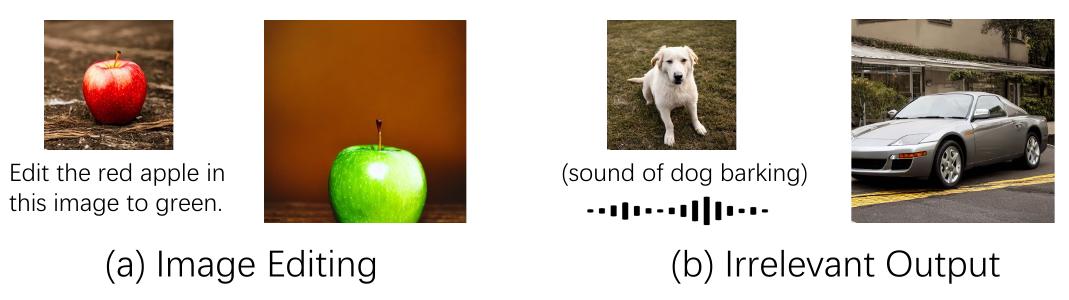}
    \caption{Failure cases of ModaVerse. (a) The model can only generate entirely new images and cannot modify the original pixels. (b) The model tends to generate irrelevant outputs in the absence of language instructions during the input phase.}
    \label{fig:failure}
\end{figure}

In exploring the capabilities of ModaVerse, Figure~\ref{fig:failure} includes some challenging scenarios where the model's performance can be further enhanced. Specifically, Example (a) illustrates a current limitation of the model in image editing tasks, where it fails to retain the original background and layout of the input images. Instead of modifying the existing image, the model generates an entirely new one. This limitation highlights a specific challenge in our approach, particularly for tasks that require fidelity to the original image's resolution and details. However, this can potentially be addressed by integrating an additional editing model into the model zoo at the final response generation stage, a development we leave for future work. Another notable case is Example (b), where, in the absence of language clues at the input phase, the model tends to produce random, irrelevant outputs. This issue arises because the instruction-following trained model relies on given language instructions for reasoning out the expected response. Without such clues, it may struggle to produce appropriate responses.

\vspace{-0.1cm}
\section{Conclusion}
\vspace{-0.1cm}
In this paper, we have presented ModaVerse, a MLLM capable of interpreting and generating data in various modalities. This model diverges from existing MLLM frameworks by adopting a synergistic approach that merges adaptor training with the LLM-as-agent methodology. By employing adaptors, ModaVerse effectively aligns the text-based LLM with multi-modal inputs through a set of linear projection layers. This enhances its capability to interpret a diverse array of input modalities. On the output side, instead of training additional projection layers to align the output space with generative models, we treat the LLM as an agent. This agent produces a meta-response containing invocation details, which are then parsed to activate generative models for generating the final response. This integrative Adaptor+Agent training paradigm not only streamlines the complex multi-stage feature alignment process but also significantly boosts the efficiency of the training process, offering an alternative for the training of MLLMs. 
For future work, we aim to address the current framework's limitations and weaknesses, such as preserving the original layout information of inputs, thereby broadening its applicability to scenarios requiring original information, like image and video editing.

\end{document}